\documentclass{article}

\usepackage[final, nonatbib]{nips_2016}
\usepackage[utf8]{inputenc} % allow utf-8 input
\usepackage[T1]{fontenc}    % use 8-bit T1 fonts
\usepackage{url}            % simple URL typesetting
\usepackage{booktabs}       % professional-quality tables
\usepackage{amsfonts}       % blackboard math symbols
\usepackage{nicefrac}       % compact symbols for 1/2, etc.
\usepackage{microtype}      % microtypography
\usepackage{todonotes}      % to indicate remaining tasks
\usepackage{mlmacros}		% macros to simplify ones life
\usepackage[fleqn]{amsmath}	% fleqn for left alignment of math blocks (as in classicthesis)
\usepackage{amssymb}
\usepackage{multirow}
\usepackage{graphicx}
\usepackage{subcaption}

\title{Classification of sparsely labeled spatio-temporal data through semi-supervised adversarial learning}

\author{Atanas Mirchev\\ Department of Computer Science\\ Technical University of Munich \\ \texttt{atanas.mirchev@tum.de} \\  \And Seyed-Ahmad Ahmadi \\ Department of Neurology \\ LMU Klinikum Grosshadern \\ \texttt{aahmadi@med.lmu.de}}

\variables{x,y,z,c,X,Z}
\probdists{p,q}

\begin{document}

\maketitle
\begin{abstract}
In recent years, Generative Adversarial Networks (GAN) have emerged as a powerful method for learning the mapping from noisy latent spaces to realistic data samples in high-dimensional space. So far, the development and application of GANs have been predominantly focused on spatial data such as images. In this project, we aim at modeling of spatio-temporal sensor data instead, i.e.\ dynamic data over time. The main goal is to encode temporal data into a global and low-dimensional latent vector that captures the dynamics of the spatio-temporal signal. To this end, we incorporate auto-regressive RNNs, Wasserstein GAN loss, spectral norm weight constraints and a semi-supervised learning scheme into InfoGAN, a method for retrieval of meaningful latents in adversarial learning. To demonstrate the modeling capability of our method, we encode full-body skeletal human motion from a large dataset representing 60 classes of daily activities, recorded in a multi-Kinect setup. Initial results indicate competitive classification performance of the learned latent representations, compared to direct CNN/RNN inference. In future work, we plan to apply this method on a related problem in the medical domain, i.e.\ on recovery of meaningful latents in gait analysis of patients with vertigo and balance disorders.
\end{abstract}

\section{Introduction}
In recent years, Generative Adversarial Networks (GANs) \cite{gans} have gained a lot of popularity as a generative modeling framework. Such networks have been successfully applied to various tasks \cite{dcgan,salimans2016improved,began,nlpgans1,policygans}, where the main body of work focuses on modeling the distributions of natural images. The goal of modeling with GANs is to approximate a target ground truth distribution as well as possible. To this end, a generator is trained to produce a model distribution by transforming a random noise seed. A discriminator (also called critic) then must try to distinguish between samples from that model distribution and from the target distribution. Both the generator and discriminator oppose each other's objective, and are alternatingly trained with the goal that a given distance between the generated model distribution and the real distribution is minimized. Since the model distribution is solely specified by the transformation of a random variable seed by the generator, its probability density function need not be known in closed form. This is the main theoretical property of GANs that has drawn researchers' attention: GANs are a method for creating \emph{implicit models}. Thus, GANs have the potential ability to learn high-dimensional, highly structured distributions (e.g.\ natural images) and can be used to enable extensions of other theoretical frameworks (such as variational inference \cite{avb}). 

Despite being a very popular method, generative modeling with GANs has not been fully explored yet. In particular, when it comes to modeling sequential time-series data, proposed methods have just recently started to emerge \cite{seqgan,nlpgans1,nlpgans2,musicgans1,musicgans2,videogans}.
In this work we choose to study the modeling capabilities of the GAN framework, applied to spatio-temporal sensor data -- such as human action sequences. Besides evaluation of the generated samples, we additionally focus on the quality of encodings $\bz$, where $\bz$ is a \emph{global}, low-dimensional salient vector corresponding to the observed temporal sequence. However, the original GAN formulation does not explicitly provide a method for obtaining said encodings. Thus, lately a number of works \cite{infogan,bigan,ali} have proposed extensions or reformulations of the GAN objective to incorporate an \emph{encoder} network, which effectively serves as an approximation of the mapping between observed data samples (in this case sequences) $\bx$ and their latent representation $\bz$. In the context of the GAN framework, such an encoder can be trained in \emph{unsupervised} or \emph{semi-supervised} fashion. This means that available data can be utilized even if it is not labeled. It is thus of interest whether the encodings obtained that way can lead to improved classification (or regression) performance on unseen data, compared to a regular discriminative model that has only been trained on the labeled portion of the data set. This is of particular importance for data sets where a very small amount of the data is labeled (e.g. medical domain).

In the following section, the differences between discriminative and generative modeling are illustrated. Then related work regarding generative modeling of sequences with GANs is presented. Section 4 describes the possibilities of extending the GAN framework to obtain latent encodings, i.e.\ an approximation of the posterior $\p{\bz}{\bx}$. Section 5 introduces the data set used for experiments. In section 6, our specific implementation and architecture is discussed in more detail and section 7 presents the experimental results for the chosen model.

\section{Discriminative and Generative Models}
Traditionally, \emph{discriminative models} approximate the distribution $p(\mathbf{y} | \bx)$ where $\mathbf{y}$ denotes the label (class or continuous value) and $\bx$ the sample for which predictions should be made. With the increasing popularity of deep learning, nowadays deep discriminative models are applied to solve various practical tasks in different domains. A few examples are classification of images \cite{2012imagenet,vgg,googlenet,resnet,bigdataCV}, speech recognition \cite{2012speech,gravesspeech,speechSotA} or action recognition \cite{actionSotA}. The success of deep models can be arguably attributed to properties like the increased representational capacity (universal function approximators \cite{universalApprx}), the increased flexibility (not requiring hand-crafted features, able to work with large amounts of raw data) and the possibility to utilize variations of SGD (stochastic gradient descent \cite{sgd}) to process very large data sets. Therefore, deep discriminative models excel at making predictions when abundant amounts of labeled data are present. However, when this is not the case and label acquisition is costly (e.g.\ medical domain), the discriminative approach is limited by the amount of available labels, and thus demands some form of regularization to account for the imperfectly represented ground truth.

In contrast, \emph{generative models} focus on modeling the evidence of the samples directly: $\p{\bx}$, disregarding the label information at first. On its own, this definition represents fully unsupervised learning of latent salient structures present in the samples $\bx$. The benefit is that such models are not restricted by the amount of labels available and can better utilize sparsely-labeled data sets, a setting which can be problematic for discriminative models. Moreover, a central topic in generative modeling is obtaining an approximation of the true posterior $\p{\bz}{\bx}$. The latent representation $\bz$ of a sample $\bx$ should account for all regularities and structure in the sample, which would make it a good feature representation for further classification and regression tasks. An approximation of the posterior $\p{\bz}{\bx}$ can be obtained by explicitly integrating it in the training of aforementioned generative models. Two prominent generative frameworks are GANs and VAEs (Variational Auto-encoders) \cite{vaes}. In both cases deep neural networks can be utilized, increasing the modeling capacity of the overall generative model and allowing it to approximate complex real-world data distributions.

In this work, we aim to explore how well GANs can be utilized as a generative model for a large human activity data set (temporal data), when different amounts of the available sequences are labeled.

\section{Related Work}
\paragraph{Generative models}
In recent years, deep generative models have become the de facto standard for unsupervised learning of real-world data of various modality. Three prominent examples are autoregressive models \cite{pixelrnn,wavenet}, Variational Autoencoders (VAEs) \cite{vaes} and Generative Adversarial Networks (GANs) \cite{gans}. 

In auto-regressive models, such as PixelRNN \cite{pixelrnn}, the contextual dependencies between spatio- or temporaly-proximate parts of a given data sample are learned by a generating network. Thus, the model learns how to generate a realistic looking segment $\bx_t$ of the data sample given a set of already generated parts $\bx_{1:t-1}$ (context), such that $\p[\theta]{\bx_t}{\bx_{1:t-1}}$ is maximized w.r.t.\ the model parameters $\theta$.

Conversely, a Variational Autoencoder maximizes a lower bound (ELBO) of the log of the model evidence $\log \p[X]{\bx}$. This is done by training a generating network (decoder), defined by a set of parameters $\theta$, which realizes the mapping from latent space to data space: $\p[\theta]{\bx}{\bz}$. Simultaneously, the framework facilitates the minimization of the KL divergence between an approximate inference distribution $\q[\phi]{\bz}{\bx}$ and the true posterior $\p{\bz}{\bx}$.

As a counterpart, GANs (the focus of this work) represent an instance of deep neural samplers. The ground truth distribution of the data $\p[X]{\bx}$ is approximated by a random variable transformation of a seed distribution $\p{\bz}$, where the transformation is typically parameterized with a neural network \emph{generator} $G$: $\bx_g = G(\bz), \bz \sim \p{\bz}$. Thus, $G$ implicitly defines a probability distribution $\p[G]$ whose PDF need not have an analytical form. To train the generator, a discriminator $D$ (a neural network) is defined to disambiguate between real samples $\bx \sim \p[X]{\bx}$ and generated ones $\bx_g \sim \p[G]{\bx}$. The following adversarial (minimax) objective $\min_{\theta_G} \max_{\theta_D} \loss{\theta_G, \theta_D}$ is posed:

\begin{equation}
\min_{\theta_G} \max_{\theta_D} \loss{\theta_G, \theta_D} := \expc[\bx \sim \p[X]{.}]{\log D(\bx)} + \expc[\bx_g \sim \p[G]{.}]{\log (1 - D(\bx_g))}
\end{equation}

However, the original GAN framework does not specify a direct procedure for mapping data into latent space. A number of proposed extensions of the GAN formulation tackle this issue and are discussed in more detail in Section 4.

\paragraph{Stabilizing GAN training}
In practice, GANs in their original formulation have proven notoriously difficult to train. In \cite{wgan1} it is argued that training the discriminator $D$ to optimality (a theoretical requirement) often leads to vanishing gradients for the generator in high-dimensional data spaces, as long as the GAN is trained to minimize the Jensen–Shannon divergence. To remedy this behaviour, the authors of WGAN (Wasserstein GAN) \cite{wgan2} propose a different objective that approximately minimizes the Wasserstein distance between $\p[X]$ and $\p[G]$, keeping the adversarial setup:

\begin{equation}
\min_{\theta_G} \max_{\theta_D} \loss{\theta_G, \theta_D} := \expc[\bx \sim \p[X](.)]{D(\bx)} - \expc[\bx_g \sim \p[G](.)]{D(\bx_g)}
\end{equation}

Besides stabilizing training by allowing for $D$ to be trained to optimality, a further advantage of the WGAN objective is that the Wasserstein distance can be monitored during training and can indicate convergence. However, it is required that $D$ (here called a \emph{critic}) is K-Lipschitz continuous for all $\bx$, where K is a positive constant. To satisfy this prerequisite, previous works have proposed weight clipping of the critic parameters \cite{wgan2} or a gradient penalty regularization term \cite{wgangp}. As of recent, \cite{spectralnorm} have shown that \emph{spectral normalization} of all $D$ parameters outperforms other K-Lipschitzness enforcing methods in terms of stability and computation time.

\paragraph{Adversarial modeling of sequential data}
In the domain of unsupervised adversarial learning of sequential data, SeqGAN \cite{seqgan} has shown that GANs can be applied to the generation of short sequences of natural language using reinforcement learning. Recently, that line of work was continued in \cite{nlpgans1,nlpgans2} and the authors of \cite{nlpgans2} specifically express the need for more elaborate training heuristics when generating natural language sentences with RNN generator and RNN discriminator. This illustrates the increased difficulty of the generation task when dealing with temporal dependencies in the data, particularly for the NLP domain where features are discrete and gradients are difficult to propagate back. A maximum-likelihood augmentation of the GAN framework is proposed in \cite{nlpgansimproved} to stabilize training when discrete data (e.g. language) is modeled. Furthermore, GANs have been used for music generation in \cite{musicgans1, musicgans2}, where the sequences are also represented by discrete tokens.

To the best of our knowledge, existing works which model temporal sequences with adversarial learning predominantly target discrete data, such as natural language. The focus so far has usually been on the generated samples' quality and not on the corresponding latent space. Two main contributions of this work are thus the application of a GAN to a continuous-valued action recognition data set, and the evaluation of the obtained \emph{latent encodings}.

\section{Obtaining Encodings from a Generative Adversarial Model}
In generative modeling it is often desirable to obtain salient latent encodings of the modeled data. The hope is that the encodings could capture the structure (semantics) present in the observed data samples. For example, this could be the inherent clustering in the data set (w.r.t.\ the different classes), which can be useful for further classification tasks. One of the objectives of this work is to explore the representative power of encodings obtained through adversarial learning  of spatio-temporal sequences.

In their original formulation, GANs specify a generation process that transforms a random noise seed $\bz \sim \p{\bz}$ into a generated data sample $\bx_g \sim \p[G]{\bz}$. However, a reverse mapping going from data space $\bX$ to the space of latent encodings $\bZ$ is not provided, and requires an extension of the GAN framework. As of recent, a number of solutions to this problem have been proposed. In the following, $G$ denotes the generator, $E$ denotes the encoder and $D$ denotes the discriminator (critic).

\begin{figure}
	\centering
	\includegraphics[width=0.8\textwidth]{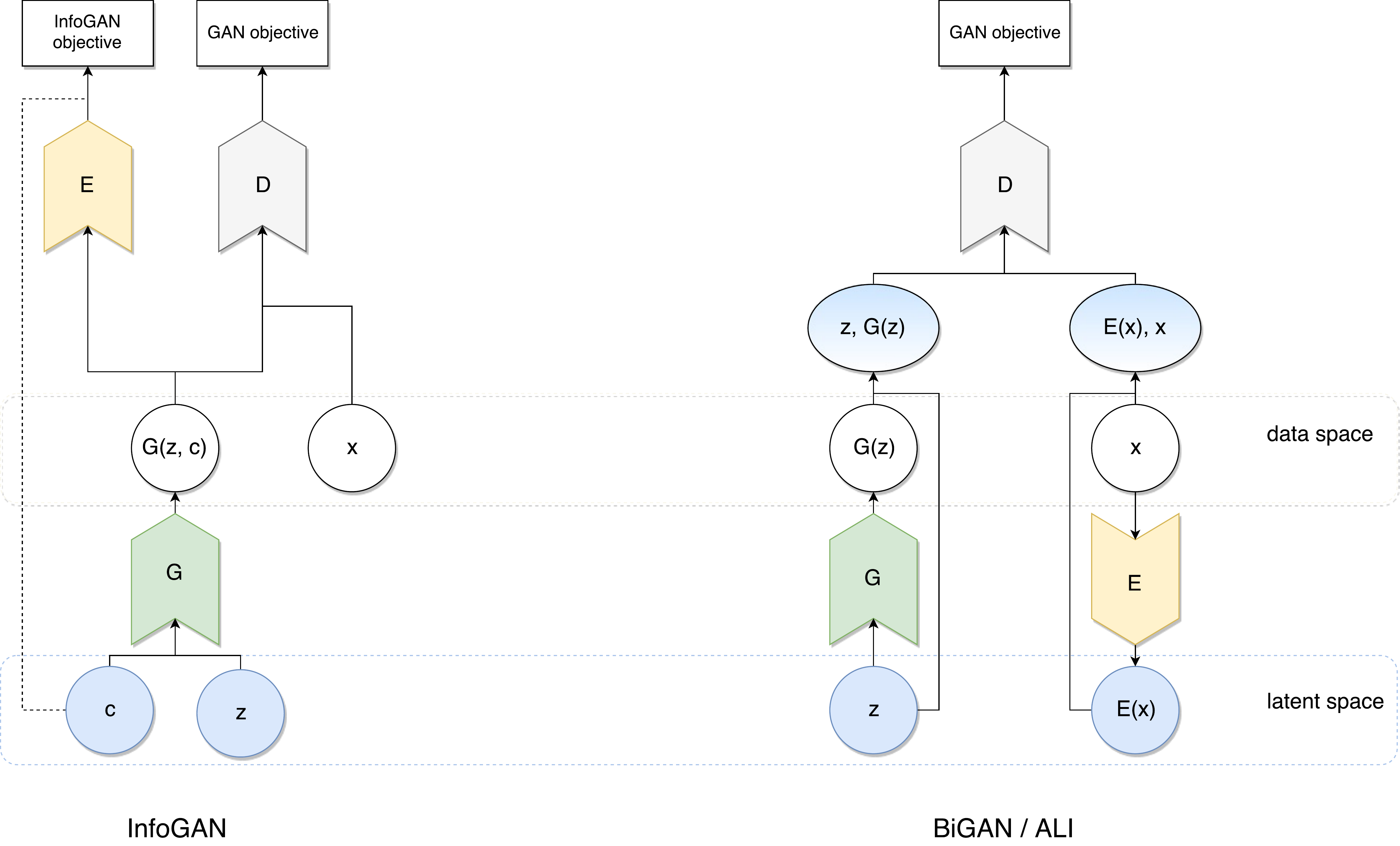}
	\caption{Comparison of two different architectures for obtaining latent encodings in the adversarial setting. \emph{Left:} architecture of InfoGAN; \emph{Right:} architecture of BiGAN and ALI;}
	\label{fig:gans}
\end{figure}

\subsection{InfoGAN}
InfoGAN \cite{infogan} is an information-theoretic extension of the GAN framework. It splits the latent code (\emph{GAN seed}) into a set of salient latents $\bc$ and a noise vector $\bz$. It further adds an additional regularization term to the GAN objective, which maximizes a lower bound of the mutual information $I(\bc; G(\bc, \bz))$ between salient latents $\bc$ and the data sample $\bx_g = G(\bc, \bz)$ generated from both $\bc$ and $\bz$:

\begin{equation}
	\loss[I]{\theta_G, \theta_E} = \expc[\bc \sim \p[\bc]{.}, \bx_g \sim \p[G]{.}{\bz, \bc}]{\log \p[E]{\bc}{\bx_g}} + H(\bc) \le I(\bc; \underbrace{G(\bc, \bz)}_{\bx_g})
\end{equation}

Intuitively, maximizing the mutual information $I(\bc; G(\bc, \bz))$ w.r.t.\ the \emph{encoder} and \emph{generator} in parallel forces the salient portion of the latent codes to capture integral information about the generated samples $\bx_g$. The InfoGAN loss term drives both the generation process and the training of the encoder because it is maximized w.r.t.\ both $\theta_G$ and $\theta_E$. Furthermore, the authors of InfoGAN show that the encoder distribution $\p[E]{\bc}{\bx}$ converges to the true posterior:

\begin{equation}
	\expc[\bx]{\kl{\p[E]{\bc}{\bx}}{\p{\bc}{\bx}}} \rightarrow 0
\end{equation}

Thus, the encoder network acts as an approximator to the mapping from data space $\bX$ to latent space $\bZ$.
This is a useful property, because it lets us obtain latent encodings for any sample from the empirical data set, which can then be further used for e.g.\ classification tasks after the generative model has been trained. An obvious prerequisite for the successful application of this approach is that the generator $G$ has been trained to approximate the empirical data distribution $\p[X]$ well. The InfoGAN architecture is summarized in Figure~\ref{fig:gans}, left.

\subsection{BiGAN \& ALI}
One alternative to the InfoGAN formulation is the approach in BiGAN and ALI \cite{bigan,ali}. This approach generatively models the complete joint distribution $\p{\bx, \bz}$ by defining the min-max objective as:

\begin{multline}
	\min_{\theta_G} \min_{\theta_E} \max_{\theta_D} \loss{\theta_D, \theta_E, \theta_G} = \expc[\bx \sim \p[X]{.}]{\expc[\bz \sim \p[E]{.}{\bx}]{\log D(\bx, \bz)}} + \\ \expc[\bz \sim \p[Z]{.}]{\expc[\bx \sim \p[G]{.}{\bz}]{\log (1 - D(\bx, \bz))}}
\end{multline}

An encoder realizes the distribution $\p[E]{\bz}{\bx}$ which approximates the true posterior $\p{\bz}{\bx}$. Note that in BiGAN the encoder is applied to real samples from the data empirical $\p[X]{\bx}$, and not to generated samples $\bx_g \sim G(\bz)$. The discriminator in BiGAN needs to distinguish between $(\bx,\bz)$-pairs of the following two types: either $\bx$ is a real sample and $\bz \sim \p[E]{\bz}{\bx}$ is its respective encoding, or $\bx \sim \p[G]{\bx}{\bz}$ is a generated sample obtained by transforming $\bz \sim \p{\bz}$ (the noise prior seed). Another important difference to InfoGAN is that the encoder $E$ must map to the whole latent space, and not just a subset of salient latents. A diagram with the summarized BiGAN / ALI architecture can be found in Figure~\ref{fig:gans}, right.

While this approach intricately involes E in the overall adversarial learning procedure, it also requires that $D$ discriminates in the higher dimensional space of the complete joint distribution $\p[X, Z]$. At the same time, generating long spatio-temporal sequences is computationally intensive on its own and in the studied case requires a complex, auto-regressive RNN to transform the global encoding $\bz$ into a sequence $\bx$ (see section 6). Because of practical considerations in terms of computational resources, we thus opt for the method proposed in InfoGAN.

\section{Data Set}
To evaluate the performance of our model, we use the NTU RGB+ D \cite{ntu} large scale action recognition data set. For our experiments only the \emph{joints} modality of the data set is used - a sequence is represented by the 3D coordinates of 25 skeleton joints at each time step. We split the data into a training set consisting of 20 subjects (40,320 samples in total) and a test set consisting of the remaining 20 subjects (16,560), which is the exact split used in the original paper. We pre-process the skeleton sequences by sub-sampling every second frame, resulting in a maximal sequence length of 150 steps. We further restrict our data to single subject actions only, to avoid padding the sequence frames with filler values (which would be necessary if 2- or 3-subject actions were present). This leaves us with 31772 samples in the training set and 13115 samples in the test set. Additionally, the skeletons are centered in every frame, with the hip joint placed at the origin. 

\section{Proposed Method}

\begin{table}[t]
  \centering
  \begin{tabular}{lll}
    \toprule
    \multicolumn{3}{c}{Wasserstein InfoGAN} \\
    \midrule
    \multirow{2}{*}{\textbf{Generator (G)}} & \textbf{layers} & 1 x AR-RNN(hidden layers=1, units=64) \\
    & \textbf{activation} & softsign \\
    & \textbf{learn. rate} & 1e-4 \\
    \midrule
    \multirow{2}{*}{\textbf{Critic (D)}} & \textbf{layers} & 2 x Conv1D(filters=64, kernel size=5)*, 1 x GlobalMaxPooling* \\
    &  & 1 x Dense(units=64) \\
    & \textbf{activation} & tanh \\
    & \textbf{learn. rate} & 1e-4 \\
    \midrule
    \multirow{2}{*}{\textbf{Encoder (E)}} & \textbf{layers} & 2 x Conv1D(filters=64, kernel size=5)*, 1 x GlobalMaxPooling* \\
    &  & 1 x Dense(units=64) \\
    & \textbf{activation} & tanh \\
    & \textbf{learn. rate} & 1e-4 \\
    \midrule
    \multicolumn{3}{c}{CNN classifier} \\
    \midrule
      & \textbf{layers} & 2 x Conv1D(filters=64, kernel size=5), 1 x GlobalMaxPooling \\
    &  & 1 x Dense(units=64) \\
    & \textbf{activation} & tanh \\
    & \textbf{learn. rate} & 1e-4 \\
    & \textbf{dropout} & 0.1 \\
    \bottomrule
    & & \\
  \end{tabular}
  \caption{Architecture specification of the proposed InfoGAN model and the CNN classifier baseline. First 3 rows describe G, D and E of InfoGAN. AR-RNN stands for Autoregressive-RNN. Last row describes the CNN classifier. Note that the layers marked with * are shared between E and D, for computational efficiency.}
  \label{architecturetable}
\end{table}
\paragraph{Generative model} We choose to follow the InfoGAN framework. The implemented GAN consists of three deep neural network models: G (the generator), D (the critic) and E (the encoder). Table \ref{architecturetable} contains the details about the hyper-parameters of the three models. The encoder E and critic D share the weights of the 1-D convolutional layers, which positively affects training times.

The input for the generator G is formed by sampling from the salient and noise latent seeds: $\bc \sim \p{\bc}, \bz \sim \p{\bz}$. Note that those latent variables are \textbf{global} for the whole sequence, as we are interested in features describing the sequence as a whole (such as class indication, etc.).  We choose to model $\bc$ as a categorical random variable with 60 classes. This is done in order to later explore whether the model was able to capture the inherent clustering w.r.t. the different action classes in the salient latent $\bc$. The additional noise seed $\bz$ is sampled from a 64-dimensional uniform distribution: $U([0, 1]^{64})$. The sampled global latents are repeatedly fed as input to G at every time step of the sequence generation. We found that a regular RNN generator has difficulties converging to a meaningful solution. Thus, we implement the generator as a more powerful Autoregressive-RNN, where the output from the previous time step $\by_{t-1}$ is fed as an input for the current time step. The resulting formula for one step of the RNN model is:

\begin{equation}
\by_{t} = RNNCell(\by_{t-1}, \bz, \bc)
\end{equation}

where RNNCell denotes a GRU \cite{gru} in our case.

Note that in contrast to models like PixelRNN \cite{pixelrnn} or WaveNet \cite{wavenet}, where the \emph{ground truth} from the previous time step is fed as input to the current step (also known as \emph{Teacher Forcing}), in our setup we can only feed the previous \emph{prediction}. Feeding a ground truth to G would be wrong, as the salient latents $\bc$ can only be sampled independently from the fed sequence $\bx_{real}$. However, $\bc$ will need to explain the generated sequence $\bx_g$ well (due to the InfoGAN mutual information loss). If $\bx_{real}$ is used as ground truth, that would tie it to $\bx_g$, breaking the assumption that $\bc$ is sampled independently from $\bx_{real}$.

Furthermore, in order to generate actions of different duration, we sample a desired length $l$ from a prior distribution on the sequence lengths for every sequence that needs to be generated. The output of the auto-regressive RNN is then masked based on the sample length, so that the outputs $\{\bx_{g,t}\}_{t > l}$ are zeroed out and thus are ignored by the discriminator and encoder. The length prior model we used was empirically fit to the sequence lengths in the data set (ranging from 0 to 150 steps), and was set to $min\{20 + Beta(\alpha=12.5, \beta=2.5), 150\}$, where $Beta(.)$ denotes the Beta distribution.

\begin{figure}
	\centering
	\includegraphics[width=1.0\textwidth]{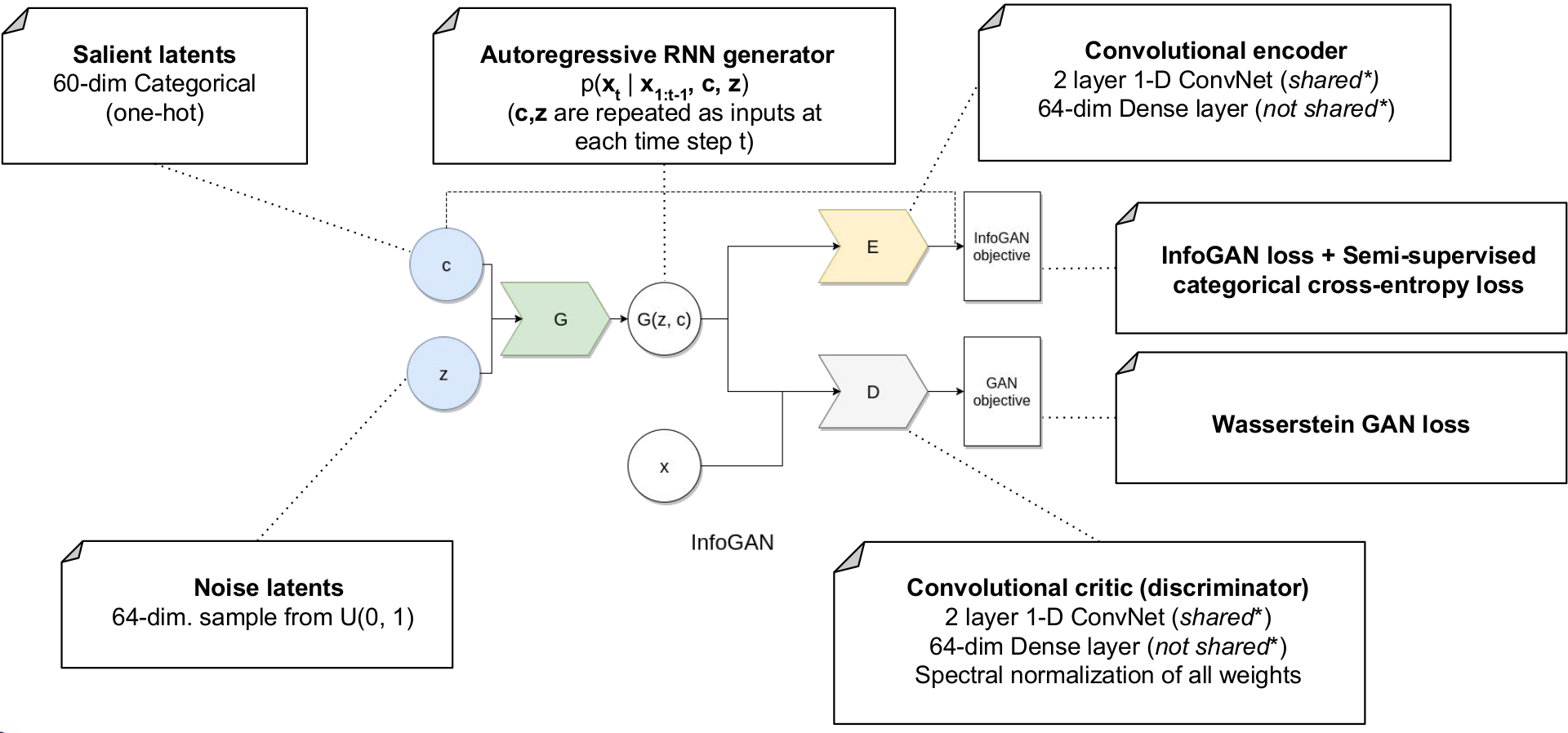}
	\caption{Diagram of the proposed method's architecture.}
	\label{fig:architecture}
\end{figure}

A summary of the overall architecture of the proposed method is presented in Figure \ref{fig:architecture}.

\newpage

\begin{figure}[h]
	\centering
	\includegraphics[width=1.0\textwidth]{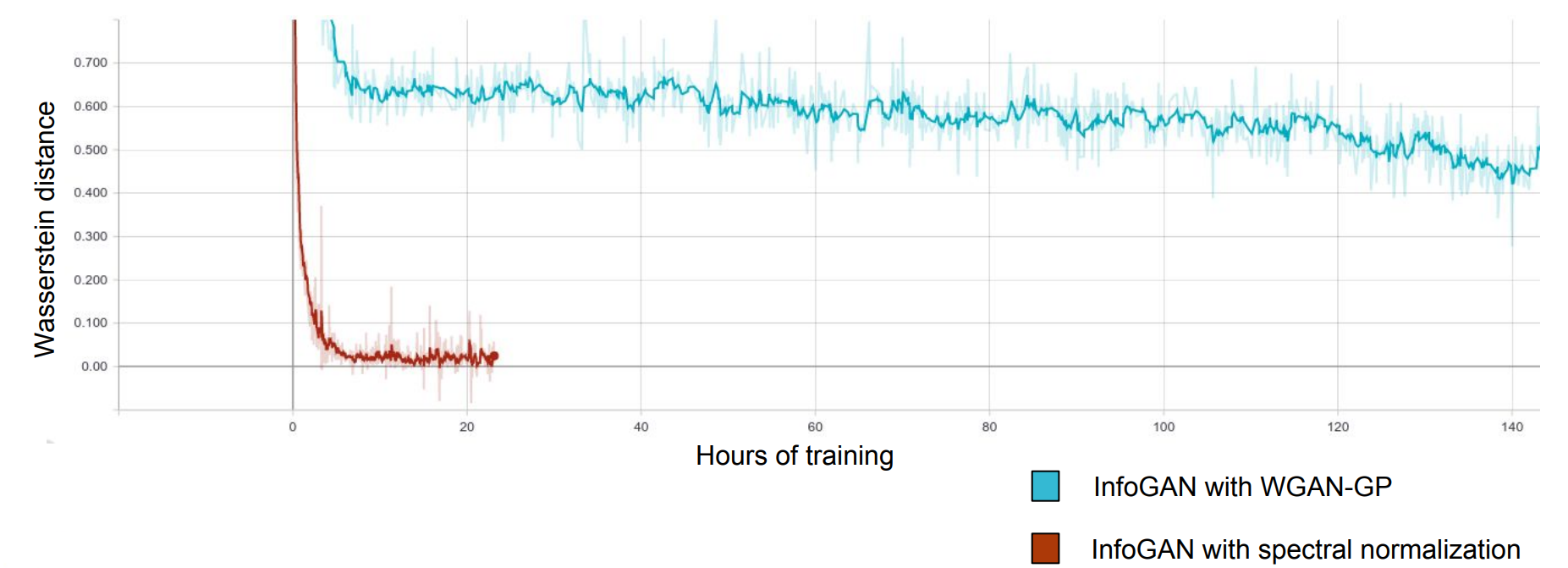}
	\caption{Comparison between the training time with spectral normalization \cite{spectralnorm} and the training time with WGAN-GP \cite{wgangp}.}
	\label{fig:loss-curves}
\end{figure}

To stabilize training, the Wasserstein loss \cite{wgan2} was used instead of the original GAN loss. Convergence during training is indicated by the Wasserstein metric saturating at a positive value or reaching 0 (cf. Figure \ref{fig:loss-curves}). Since the Wasserstein formulation requires a K-Lipschitz critic D, we employ the \emph{spectral normalization} proposed in \cite{spectralnorm}. We find that spectral normalization is necessary to stabilize the training of InfoGAN when deeper neural networks are utilized (more than 2 layers), whereas WGAN-GP \cite{wgangp} fails to do so in our setup. Furthermore, the direct normalization of the weights allows one to use higher learning rates without instability in the gradients flowing through the network. This is not easily possible with WGAN-GP, where the gradient penalty is quite sensitive to the chosen learning rate. With this observation in mind, spectral normalization of the discriminator weights allowed us to train up to 10 times faster compared to WGAN-GP (cf. Figure~\ref{fig:loss-curves}).

\paragraph{Classifier baselines} In order to evaluate the quality of the latent encodings $\bc$ obtained through semi-supervised GAN learning for further classification tasks, first some baselines need to be set. To that end, we define a straightforward 1-D CNN classifier network (architecture details in Table \ref{architecturetable}, last row), whose performance will be compared to the encoder network trained in the GAN framework. For a fair comparison, the architectures, and in particular the capacity, of both the CNN classifier and encoder E are kept the same.

\section{Experiments}
We ran experiments with varied amounts of labels from the NTU RGB+ D training set: 5\%, 10\%, 20\%, 40\% and 100\%, in order to simulate real-world scenarios where labeled data might be scarce. The CNN classifier is trained in a fully \textbf{supervised} fashion on the provided labeled portion of the data set. The encoder E is trained following the InfoGAN objective on the full data set, as no labels are needed. An additional supervised loss \footnote{For the NTU RGB+ D data set, the semi-supervised loss added to the encoder is the categorical cross-entropy of the predicted encoding $\p{\bc}{\bx_{real}}$ w.r.t. the labels of $\bx_{real}$.} is added to E and is evaluated (and errors from it are back-propagated) only for the labeled portion of training data. Thus the encoder is trained in a \textbf{semi-supervised} fashion.

\begin{figure}
    \centering
    \begin{subfigure}[b]{0.35\textwidth}
        \includegraphics[width=\textwidth]{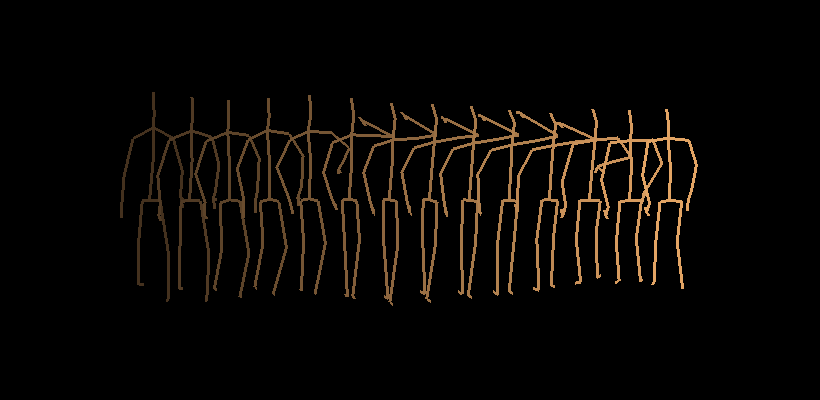}
        \caption{A salute}
        \label{fig:gen-skeleton-solute}
    \end{subfigure}
    ~ %add desired spacing between images, e. g. ~, \quad, \qquad, \hfill etc. 
    \begin{subfigure}[b]{0.282\textwidth}
        \includegraphics[width=\textwidth]{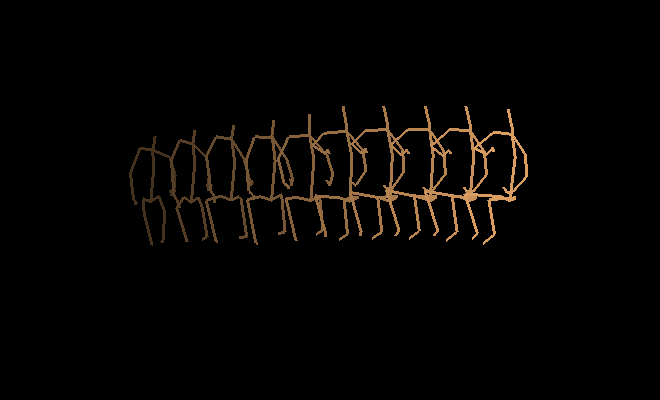}
        \caption{A sitting action}
        \label{fig:gen-skeleton-sit}
    \end{subfigure}
    \begin{subfigure}[b]{0.265\textwidth}
        \includegraphics[width=\textwidth]{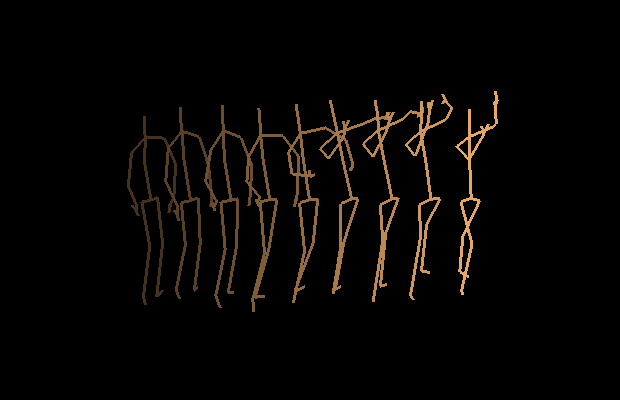}
        \caption{Raising both hands}
        \label{fig:gen-skeleton-raise-hand}
    \end{subfigure}
    
    \begin{subfigure}[b]{0.38\textwidth}
        \includegraphics[width=\textwidth]{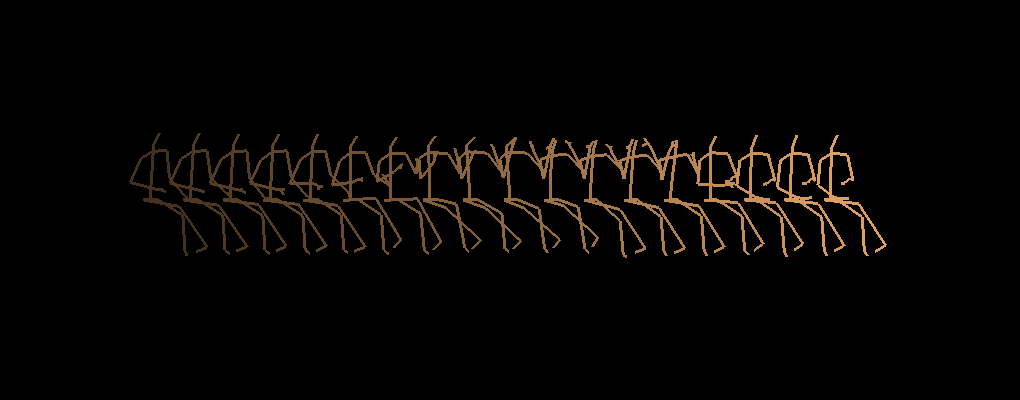}
        \caption{Putting glasses on}
        \label{fig:val-skeleton-glasses}
    \end{subfigure}
    \begin{subfigure}[b]{0.288\textwidth}
        \includegraphics[width=\textwidth]{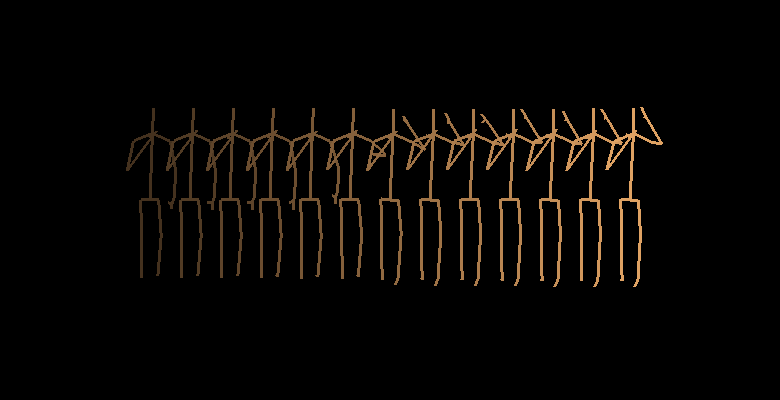}
        \caption{Raising left hand}
        \label{fig:val-skeleton-hand}
    \end{subfigure}
    \begin{subfigure}[b]{0.275\textwidth}
        \includegraphics[width=\textwidth]{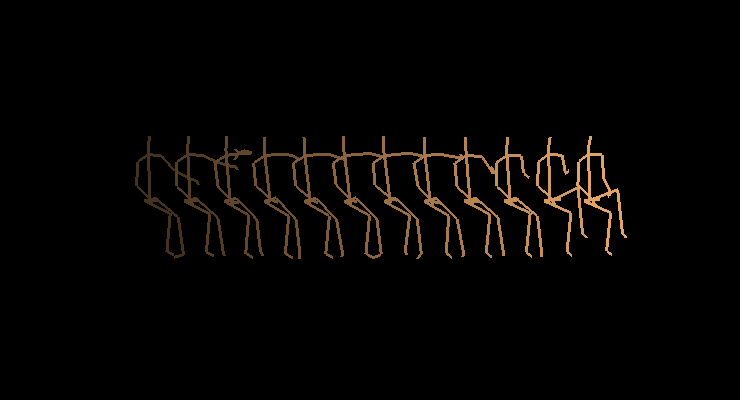}
        \caption{Sitting down}
        \label{fig:val-skeleton-sit}
    \end{subfigure}
    \caption{Visualization of example skeleton frame sequences. First row (\ref{fig:gen-skeleton-solute}, \ref{fig:gen-skeleton-sit}, \ref{fig:gen-skeleton-raise-hand}) contains generated sequences. Second row (\ref{fig:val-skeleton-glasses}, \ref{fig:val-skeleton-hand}, \ref{fig:val-skeleton-sit}) contains sequences from the NTU-RGB+ data set.}\label{fig:skeletons}
\end{figure}

For all levels of supervision, the proposed GAN was able to generate meaningful skeleton action sequences that qualitatively resemble the ground truth actions from the data set. Figure \ref{fig:skeletons} shows 3 generated activity sequences (first row) and 3 real activity sequences from the test set (second row). The movements in the generated sequences appear smooth and realistic, showing that the chosen autoregressive RNN generator was able to learn the spatio-temporal dynamics. The generated actions are varied and of different length (roughly indicated by the number of skeleton "shadows" in Figure \ref{fig:skeletons}), as is the case in the original data set. Furthermore, Figure \ref{fig:gen-skeleton-solute} exhibits an example of a generated sequence where the action is immediately identifiable. It is also worth noting that the generative model learned to ignore most of the jittery motion present in some of the real samples, which is unavoidably there due to errors in the Kinect sensors used to record the skeleton joint coordinates.

\begin{figure}
    \centering
    \begin{subfigure}[b]{0.45\textwidth}
        \includegraphics[width=\textwidth]{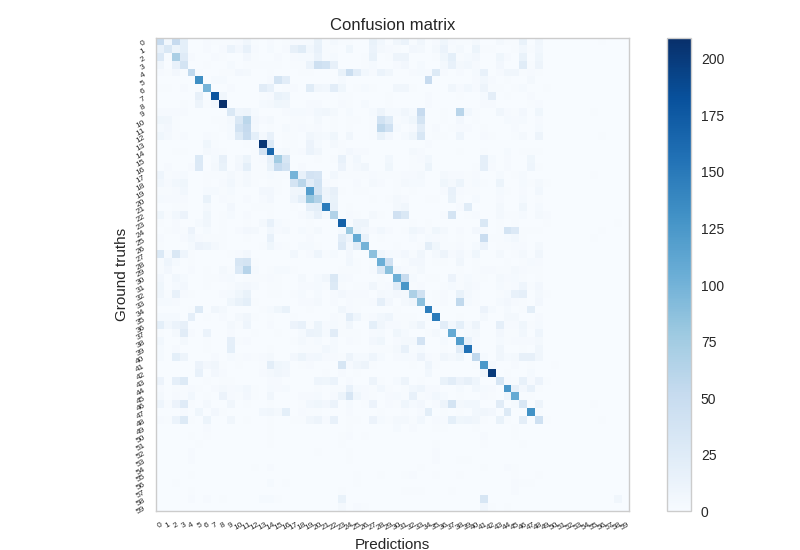}
        \caption{Confusion matrix, 5\% supervision}
        \label{fig:cm-0-5}
    \end{subfigure}
    \begin{subfigure}[b]{0.45\textwidth}
        \includegraphics[width=\textwidth]{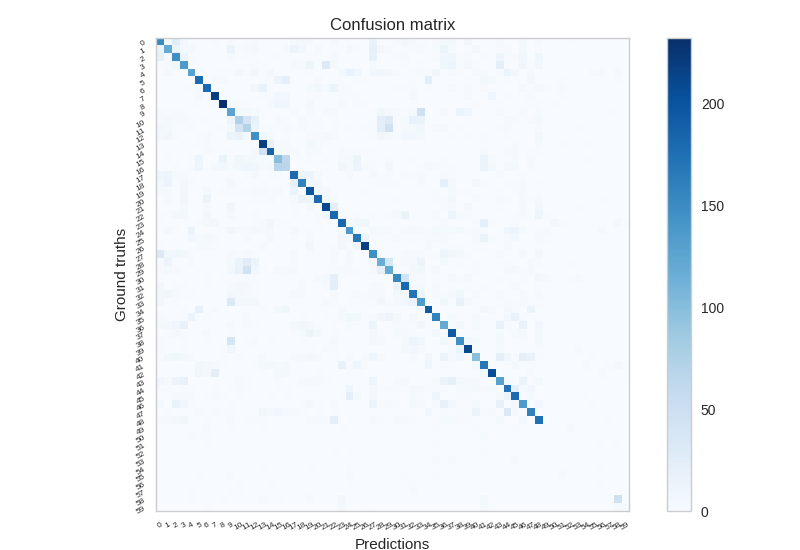}
        \caption{Confusion matrix, full supervision}
        \label{fig:cm-1-0}
    \end{subfigure}
    \caption{Confusion matrices for classification based on the obtained InfoGAN latent encodings, for the lowest level of supervision at 5\% (\ref{fig:cm-0-5}), and for the highest level of supervision (fully supervised, \ref{fig:cm-1-0}). Note that the lower right corner (last 10 classes) should be empty, as we train on single-subject frames only and the last 10 classes correspond to multi-subject classes.}\label{fig:cms}
\end{figure}

The second aspect of generative adversarial modeling we investigate is the saliency of the inferred latents by the encoder network E (part of InfoGAN). We compare the classification accuracy scores based on the latent encodings $\bc$ (obtained through InfoGAN) to the classification accuracy scores obtained with the direct 1-D CNN classifier. The results are presented in Table \ref{classification}. For completeness, we added a column for a 2-layer GRU classifier trained directly on the original representations. From the table it is apparent that the accuracies obtained with the InfoGAN encodings are exactly on par with the accuracies from the CNN classifier and those of the RNN classifier. This holds for all levels of supervision (ranging from 5\% to 100 \%). The proposed generative model was thus able to learn the mapping from latent space to data space, such that the latent codes $\bc$ carry information about the class association of the action sequences. However, the slight advantage of the InfoGAN encodings at low supervision levels compared to the 1-D CNN classifier (e.g. 36.05 \% vs. 35.12 \%, 44.40 \% vs. 43.52 \%) is not statistically significant to conclude that training in a semi-supervised way leads to a major improvement in generalization capabilities compared to only using the labeled portion of the data set. One final observation is that the classification accuracy improves as the amount of labeled data increases, which is to be expected. This is consistently the case in the semi-supervised InfoGAN setup as well: an example comparison of the confusion matrices for the classification based on encodings with 5 \% labeled data and 100 \% labeled data is visualized in Figure \ref{fig:cms}.

\begin{table}[t]
  \caption{Classification results using the original sequences and a direct classifier or their respective encodings, obtained through InfoGAN. PA-LSTM is the original method proposed in \cite{ntu}.}
  \label{classification}
  \centering
  \begin{tabular}{ccccc}
    \toprule
    \multicolumn{5}{c}{Classification accuracies on test data}                   \\
    \midrule
    \textbf{Supervision level (\%)} & \multicolumn{4}{c}{\textbf{Accuracy (\%)}} \\
    \midrule
    & InfoGAN & CNN classifier & RNN classifier & PA-LSTM \\
    \midrule
    5\% & \textbf{36.05} & 35.12  & 35.88 & - \\
   	10\% & 44.40 & 43.52 & \textbf{44.90} & - \\
    20\% & 51.25 & 51.01 & \textbf{52.25} & - \\
    40\% & 58.55 & 58.62 & \textbf{58.85} & - \\
    100\% & 64.61 & 65.08 & \textbf{66.77} & 62.93 \\
    \bottomrule
  \end{tabular}
\end{table}

\section{Conclusion}
In this work, we show that GANs can be effectively applied to model spatio-temporal sequences of high-dimensional continuous signals with varying length. The produced samples are varied and visually realistic. We further illustrate the necessity of proper normalization of the weights of the discriminator to ensure convergence, employing state-of-the-art improvements of the GAN framework. Finally, it is shown that a global latent encoding can be used as a seed for generating the whole temporal sequence. The saliency of the learned latent representations is explored, indicating competitive performance compared to direct CNN/RNN models in terms of classification.

In future work, we plan to apply this method to data sets stemming from the medical domain. BiGAN/ALI \cite{bigan,ali} are a viable alternative to InfoGAN for obtaining latent encodings in the adversarial setup and need to be studied as well, particularly in the case where small amounts of labels are available.

\bibliographystyle{plain}
\bibliography{bibliography}

\end{document}